\documentclass[10pt,twocolumn]{article}

\usepackage[utf8]{inputenc}
\usepackage{lmodern}
\usepackage[T1]{fontenc}
\usepackage[margin=0.82in]{geometry}
\usepackage{booktabs}
\usepackage{graphicx}
\usepackage{amsmath}
\usepackage{amssymb}
\usepackage{microtype}
\usepackage{xcolor}
\usepackage[numbers,sort&compress]{natbib}
\usepackage[colorlinks=true,allcolors=blue!55!black]{hyperref}
\usepackage{caption}
\newcommand{\oxor}{\ensuremath{\oplus}}
\graphicspath{{./}}

\title{\bfseries Training-Free Lexical--Dense Fusion\\
for Conversational-Memory Retrieval}

\author{Christian Lysenst{\o}en\thanks{Licensed CC BY 4.0.}\\[2pt]
\small Inland Norway University of Applied Sciences\\
\small Visiting student, University of California, Berkeley\\
\small \texttt{christian@lysenstoen.net}\\[2pt]
\small \url{https://github.com/Chrislysen/opsem}}
\date{June 2026}

\begin{document}
\maketitle

\begin{abstract}
\noindent
Retrieving the few past turns that answer a new query, across long
multi-session histories, is the retrieval bottleneck behind long-term
conversational memory (LoCoMo, LongMemEval). Recent concurrent work,
\emph{Nano-Memory}, shows that scoring a session by the maximum query--turn
similarity (``Turn Isolation Retrieval'', late interaction) beats mean-pooled
session embeddings, attributing the effect to a \emph{Signal Sparsity Effect}.
We do \emph{not} claim that effect; we replicate it and ask what a
\emph{training-free, CPU-only} retrieval stage should add around it. We report
four findings plus an analysis of when each helps. \textbf{(1)~Fuse:} score-level fusion of the late-interaction
dense score with BM25, under a single leave-one-conversation-out weight, adds
\textbf{+8.8 to +17.2 points} of LoCoMo Hit@1 \emph{over late interaction
alone} across six encoders (all $p<10^{-4}$), reaching Hit@1 $0.752$ / NDCG@5
$0.829$ (e5-large-v2), $+11.2$\,pp over BM25. \textbf{(2)~An off-the-shelf
reranker hurts here:} a web-search cross-encoder over the fused top-10
\emph{degrades} Hit@1 by $6.9$\,pp (scoped to one reranker, one config). \textbf{(3)~Pooling matters beyond max:} a
controlled pooling-operator ablation shows top-$k$ late interaction matches
max-similarity, but a naive smooth-max (log-sum-exp) collapses
catastrophically for half the encoders---so the late-interaction family is not
uniformly safe. \textbf{(4)~Robustness and boundary:} the late$-$early gap is
large for all six encoders ($+13.5$ to $+23.7$\,pp) and tends to be larger for
larger encoders, while the marginal fusion gain shrinks; on LongMemEval-S, a
lexical regime where BM25 saturates, the net fusion gain over BM25 is small and
not significant. A per-category analysis frames the gain as a \emph{division of
labor}: dense late interaction helps most on multi-hop and temporal questions
but trails BM25 on adversarial ones, so fusion's value is combining whichever
signal a query needs. The contribution is a controlled, reproducible account
of a strong training-free retrieval recipe---not the late-interaction
retriever itself, which is Nano-Memory's. We make no claim to a complete
memory architecture; this is a retrieval-stage study.
\end{abstract}

\section{Introduction}
A long-term conversational assistant accumulates hundreds of dialogue turns
across many sessions. When the user asks a new question---``what did I say was
wrong with my new car?''---the system must retrieve the small set of past
turns that contain the answer before any reader or generator can use them.
Benchmarks such as LoCoMo \citep{locomo} and LongMemEval \citep{longmemeval}
formalize this task, and a large body of recent work studies how to index and
retrieve conversational memory \citep{secom,memgpt,zep}.

Much of that work is framed around \emph{granularity}: should the atomic
retrievable unit be a turn, a session, a summary, or a learned segment? The
literature both establishes that the choice matters and disagrees on its
direction. We argue that ``granularity'' silently bundles two independent
design axes:
\begin{itemize}\itemsep1pt
  \item the \textbf{retrieval unit}---what text span is the atomic retrievable
  item (turn, session, segment); and
  \item the \textbf{interaction function}---how a candidate's score is
  computed from its constituent turn embeddings: \emph{early interaction}
  pools the turns into one vector and then compares to the query, whereas
  \emph{late interaction} compares the query to each turn and aggregates the
  resulting per-turn scores.
\end{itemize}
A pipeline that retrieves ``sessions'' but feeds whole-session \emph{text} to
an LLM reader performs no embedding pooling and pays no early-interaction
penalty; a pipeline that \emph{embeds} a session for dense retrieval
mean-pools it and does. Both call their unit a ``session''; they differ on the
hidden axis.

\paragraph{This paper is novelty-sensitive.}
The interaction-function insight is \emph{not} ours. Concurrent work,
\emph{Nano-Memory} \citep{nanomemory}, introduces Turn Isolation Retrieval
(TIR)---scoring a session by the maximum query--turn cosine---and shows it
beats mean-pooled session embeddings, an effect it calls the \emph{Signal
Sparsity Effect}. We replicate that result under a clean control and credit it
throughout; we do not claim it. ColBERT's MaxSim \citep{colbert} shares the
``late interaction'' name but operates at the \emph{token} level for passage
ranking, a different regime from turn-level session scoring. Our question is
deliberately narrower and complementary to Nano-Memory: \emph{once a
late-interaction dense score is available, what does a training-free,
CPU-only retrieval stage still gain---and what does it lose?}

\paragraph{Contributions.}
We hold the retrieval unit fixed at \emph{session} throughout and vary only
the interaction function and the lexical/neural components around it. We claim
five empirical contributions, each traceable to a released receipt and
none of which Nano-Memory studies (its TIR is dense-only):
\begin{enumerate}\itemsep1pt
  \item \textbf{Lexical--dense fusion adds over late interaction alone}
  (\S\ref{sec:fuse}). Score-level BM25\,\oxor{}\,late-interaction fusion, with
  a single leave-one-conversation-out weight, improves LoCoMo Hit@1 by
  $+8.8$ to $+17.2$\,pp over the late-interaction dense retriever, for every
  one of six encoders (all $p<10^{-4}$).
  \item \textbf{Reranking hurts} (\S\ref{sec:rerank}). A cross-encoder
  reranker over the fused top-10 \emph{reduces} Hit@1 by $6.9$\,pp in the
  tested setting---the reflexive ``add a reranker'' move is counterproductive
  on out-of-distribution conversational queries.
  \item \textbf{A pooling-operator ablation} (\S\ref{sec:ablation}). Top-$k$
  late interaction matches max-similarity, but a smooth-max (log-sum-exp)
  operator collapses for half the encoders---late interaction is a family, and
  not every member is safe.
  \item \textbf{Robustness and a boundary} (\S\ref{sec:scale},
  \S\ref{sec:cross}). The late$-$early gap is large for all six encoders and
  tends to be larger for larger ones, while the marginal value of fusion
  shrinks; and on LongMemEval-S the net fusion gain over BM25 is small and not
  significant, because that corpus is a lexical regime where BM25 already
  saturates.
  \item \textbf{When and why fusion helps} (\S\ref{sec:analysis}). A
  per-category analysis reveals a division of labor---dense wins on
  multi-hop/temporal reasoning, BM25 wins on adversarial queries, and fusion
  hedges both---together with a monotone length--dilution effect and an
  $\alpha$-robustness/RRF comparison.
\end{enumerate}
All results are retrieval-stage, on retrieval metrics, CPU-only, and
training-free. We do not run an LLM reader and make no end-to-end QA claim.

\section{Related Work}
\label{sec:related}
\paragraph{Conversational-memory benchmarks and systems.}
LoCoMo \citep{locomo} and LongMemEval \citep{longmemeval} provide long,
multi-session histories with gold evidence and decompose the problem into
indexing, retrieval, and reading. Production memory frameworks---MemGPT/Letta
\citep{memgpt}, Zep/Graphiti \citep{zep}, and others---add structure
(summaries, knowledge graphs, extracted facts) on top of retrieval. Where
these systems combine dense and lexical evidence, they typically fuse at the
extracted-fact or graph-element level via reciprocal rank fusion
\citep{rrf} or additive scoring, not as a max over per-turn vectors fused
with BM25; the recipe we study is therefore distinct from their defaults.

\paragraph{Granularity of the memory unit.}
That the unit matters is prior art. SeCom \citep{secom} argues turn $>$
session but proposes \emph{segment}-level units with compression; LongMemEval
reports session $\ge$ round on retrieval metrics. We do not claim the
granularity point; our control deliberately holds the unit fixed and varies
the interaction function instead.

\paragraph{Interaction functions and late interaction.}
ColBERT \citep{colbert} popularized token-level late interaction (MaxSim) for
passage ranking. The conversational-memory analogue---turn-level
max-similarity for session scoring---was recently introduced by Nano-Memory
\citep{nanomemory} as Turn Isolation Retrieval, with the Signal Sparsity Effect
as its explanation. Our work takes this as a starting point and is explicit
that the late-vs-mean result is theirs.

\paragraph{Lexical retrieval and hybrid fusion.}
BM25 \citep{bm25} remains a strong lexical baseline, especially when queries
and answers share surface form. Hybrid lexical--dense retrieval and rank
fusion \citep{rrf} are standard in IR, but their interaction with
\emph{turn-level late interaction} for conversational memory---and the
question of whether they add \emph{over} a late-interaction dense
retriever---has not, to our knowledge, been isolated on these benchmarks.
That gap is what this paper fills.

\section{Preliminaries}
\label{sec:prelim}
\paragraph{Task.}
A history is a set of sessions $\mathcal{M}=\{S_1,\dots,S_M\}$, each session a
sequence of turns $S=\langle t_1,\dots,t_{|S|}\rangle$. Given a query $q$, the
retriever returns a ranking of sessions; gold relevance is the set of sessions
containing the answer evidence. We hold the retrieved \emph{unit} at the
session throughout, so all methods solve the same ranking problem and differ
only in scoring.

\paragraph{Interaction functions.}
Let $e(\cdot)$ be a frozen bi-encoder and $\cos$ cosine similarity. We compare
four operators that score a session from its turn vectors:
\begin{align}
\text{early (mean):}\quad & s = \cos\!\big(e(q),\, \widehat{\textstyle\sum_t e(t)}\big),\\
\text{late (max-sim):}\quad & s = \max_{t\in S}\cos(e(q),e(t)),\\
\text{late (top-}k\text{):}\quad & s = \tfrac{1}{k}\!\!\sum_{t\in \mathrm{top}_k}\!\!\cos(e(q),e(t)),\\
\text{late (lse-}\beta\text{):}\quad & s = \tfrac{1}{\beta}\log\!\textstyle\sum_t e^{\beta\cos(e(q),e(t))},
\end{align}
where $\widehat{\cdot}$ denotes $L_2$-normalization and lse-$\beta$ is the
smooth-max (log-sum-exp) with $\beta{=}10$. Early interaction is the standard
``embed the session'' baseline; the late operators are interaction functions
over the \emph{same} cached turn vectors.

\paragraph{Lexical--dense fusion.}
For a query and candidate session we compute a BM25 score $s_{\mathrm{BM25}}$
and a dense score $s_{\mathrm{dense}}$ (one of the operators above),
$z$-normalize each within the candidate set, and combine
\begin{equation}
s_{\mathrm{fuse}} = \alpha\, z(s_{\mathrm{BM25}}) + (1-\alpha)\, z(s_{\mathrm{dense}}),
\label{eq:fuse}
\end{equation}
with $\alpha$ selected by leave-one-conversation-out cross-validation
(LOCO-CV): for each held-out conversation, $\alpha$ is chosen on the other
nine. We also report reciprocal rank fusion (RRF) \citep{rrf} as an
alternative combiner. No parameters are learned beyond the single scalar
$\alpha$.

\paragraph{Metrics and uncertainty.}
We report Hit@1 ($=$ Recall@1), Recall@3, Recall@5, MRR, and NDCG@5---the
field-standard retrieval set, so results are comparable to the benchmark
papers. On LoCoMo we use a \textbf{conversation-cluster bootstrap}
($n_{\mathrm{boot}}{=}4000$): resampling units are whole conversations, not
questions, which is the conservative choice given within-conversation
correlation. On LongMemEval-S we bootstrap over questions. We report two-sided
95\% CIs and one-sided bootstrap $p$-values for directional hypotheses.

\section{Experimental Setup}
\label{sec:setup}
\paragraph{Data.}
LoCoMo \citep{locomo} contributes $n{=}1978$ QA examples over 10
conversations in our evaluation; gold relevance is at the session level.
LongMemEval-S \citep{longmemeval} contributes a 150-question
retrieval-challenge subset (full multi-session haystacks); we evaluate only
the genuine full-haystack rows, not the degenerate oracle rows.

\paragraph{Encoders.}
We use six frozen CPU bi-encoders spanning 22M--335M parameters and three
families: \texttt{gte-base} \citep{gte}, \texttt{bge-base} and
\texttt{bge-large} \citep{bge}, \texttt{e5-base-v2} and \texttt{e5-large-v2}
\citep{e5}, and \texttt{mxbai-embed-large} \citep{mxbai}; the LongMemEval runs
additionally use the 22M \texttt{all-MiniLM-L6-v2} \citep{sbert} for CPU
tractability. All turn and query embeddings are computed once and cached;
every reported retrieval comparison runs from cache with no model load.

\paragraph{Implementation.}
BM25 uses default parameters ($k_1{=}1.5$, $b{=}0.75$). The cross-encoder for
reranking is \texttt{ms-marco-MiniLM-L-6-v2}. Everything runs on CPU with no
training. Scripts: \texttt{tune13\_interaction.py} (interaction-function
control and fusion), \texttt{tune13b\_fusion\_vs\_late.py} (fusion vs.\
late-alone), \texttt{tune10\_rerank.py} (cross-encoder), and
\texttt{lme\_interaction.py} (LongMemEval); figures by
\texttt{make\_figures.py}.

\section{Replication: Late vs.\ Early Interaction}
\label{sec:replicate}
We first reproduce the late-over-early result under our control: the retrieval
unit is fixed at session and the \emph{only} change is the pooling operator
over identical cached turn vectors. Table~\ref{tab:lateearly} gives Hit@1;
Table~\ref{tab:densefull} gives the full metric set.

\begin{table}[t]
\centering\small
\caption{LoCoMo dense-only Hit@1: early (mean-pool) vs.\ late (max-sim) over
identical turn vectors. All $\Delta$ significant at $p<10^{-4}$
(conversation-cluster bootstrap). BM25 baseline Hit@1 $=0.640$.}
\label{tab:lateearly}
\begin{tabular}{lccc}
\toprule
encoder & early & late & $\Delta$ (late$-$early) \\
\midrule
gte-base    & 0.384 & 0.519 & $+13.50$ \\
bge-base    & 0.386 & 0.547 & $+16.08$ \\
e5-base-v2  & 0.408 & 0.580 & $+17.29$ \\
mxbai-large & 0.405 & 0.578 & $+17.29$ \\
bge-large   & 0.403 & 0.602 & $+19.87$ \\
e5-large-v2 & 0.427 & 0.664 & $+23.71$ \\
\bottomrule
\end{tabular}
\end{table}

\begin{table*}[t]
\centering\small
\caption{LoCoMo dense-only retrieval, full metrics: early (mean-pool) vs.\
late (max-sim) interaction over identical turn vectors, six encoders.}
\label{tab:densefull}
\begin{tabular}{llccccc}
\toprule
encoder & interaction & Hit@1 & R@3 & R@5 & MRR & NDCG@5 \\
\midrule
gte-base    & early & 0.384 & 0.552 & 0.673 & 0.535 & 0.532 \\
gte-base    & late  & 0.519 & 0.671 & 0.763 & 0.648 & 0.646 \\
bge-base    & early & 0.386 & 0.571 & 0.674 & 0.542 & 0.536 \\
bge-base    & late  & 0.547 & 0.699 & 0.791 & 0.675 & 0.675 \\
e5-base-v2  & early & 0.407 & 0.574 & 0.687 & 0.557 & 0.552 \\
e5-base-v2  & late  & 0.580 & 0.744 & 0.817 & 0.705 & 0.708 \\
mxbai-large & early & 0.405 & 0.589 & 0.696 & 0.558 & 0.557 \\
mxbai-large & late  & 0.578 & 0.732 & 0.812 & 0.701 & 0.703 \\
bge-large   & early & 0.403 & 0.583 & 0.695 & 0.555 & 0.555 \\
bge-large   & late  & 0.602 & 0.759 & 0.823 & 0.721 & 0.722 \\
e5-large-v2 & early & 0.427 & 0.601 & 0.712 & 0.576 & 0.573 \\
e5-large-v2 & late  & 0.664 & 0.792 & 0.863 & 0.767 & 0.769 \\
\bottomrule
\end{tabular}
\end{table*}

Late interaction improves every metric for every encoder, with Hit@1 gains of
$+13.5$ to $+23.7$\,pp ($p<10^{-4}$). This reproduces the effect attributed by
Nano-Memory to turn isolation. We stress again that this section is a
\emph{replication}: the contribution starts in \S\ref{sec:fuse}.

\section{Robustness across encoders}
\label{sec:scale}
\begin{figure}[t]
\centering
\includegraphics[width=\linewidth]{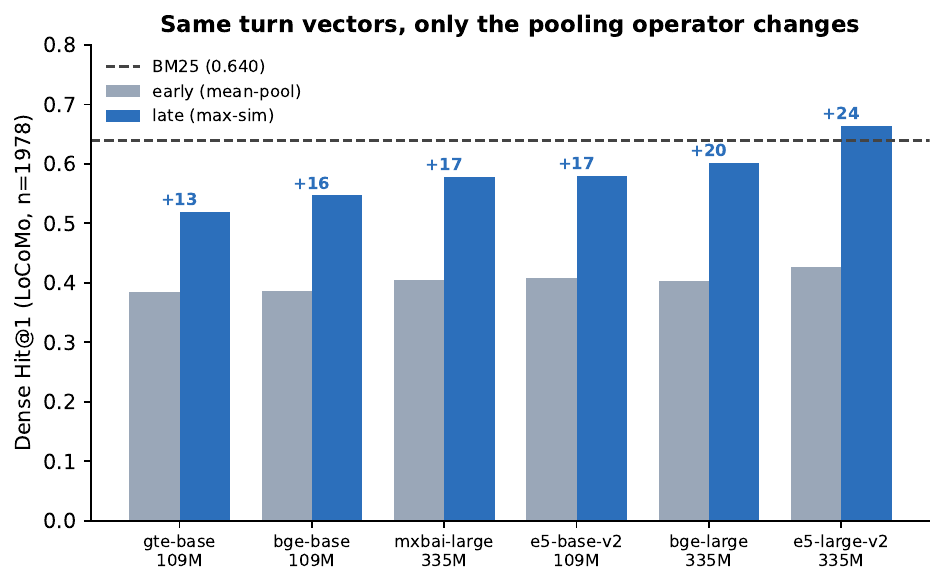}
\caption{With the retrieval unit fixed at \emph{session} and identical cached
turn vectors, switching the pooling operator from early (mean-pool) to late
(max-sim) lifts dense Hit@1 above the BM25 reference at every one of six
encoders; the $+$pp gap is annotated. Generated by \texttt{make\_figures.py}.}
\label{fig:main}
\end{figure}

The late$-$early gap is large and positive for all six encoders ($+13.5$ to
$+23.7$\,pp; Fig.~\ref{fig:main}), and it tends to be larger for larger
encoders: the three 335M-class encoders average a $+20.3$\,pp gap versus
$+15.6$\,pp for the three 109M-class ones. The trend is not monotone, however
(mxbai-large, 335M, has the same $+17.3$\,pp gap as e5-base-v2, 109M), and with
only six encoders we treat ``the gap grows with capacity'' as a soft
observation, not a law. We deliberately avoid correlating the gap against each
encoder's \emph{own} max-sim score: since mean-pool Hit@1 is nearly constant
across encoders ($0.38$--$0.43$), the gap is mechanically dominated by the
max-sim score, so such a correlation would be largely circular. A plausible
mechanism, consistent with the capacity trend but not measured directly, is
that a stronger encoder gives the answer-bearing turn a sharper similarity peak
that max-sim preserves and mean-pooling averages away. Nano-Memory likewise
reports robustness across three encoders.

\emph{Length confound.} LoCoMo gold sessions are long (the smallest length
bucket in our data is 8--15 turns), so LoCoMo alone cannot separate the
interaction effect from long-session dilution. On LongMemEval 138/150
challenge questions fall in the 8--15-turn bucket, so a length-bucketed
isolation there is also underpowered (\S\ref{sec:cross}). We therefore rest
the (suggestive) mechanism reading on the encoder-scaling trend, not on length
bucketing.

\section{Result 1: Fusion Adds Over Late Interaction Alone}
\label{sec:fuse}
Nano-Memory's TIR is \textbf{dense-only}: we verified against the paper
\citep{nanomemory} and its public repository that it uses dense retrievers
with a max-over-turns flag and no lexical, hybrid, or fusion path. The first
contribution is that adding BM25 to the late-interaction dense score helps
\emph{even though} the dense retriever is already late-interaction.
Table~\ref{tab:fuse} holds the dense term fixed at max-sim and adds
score-level BM25 fusion (Eq.~\ref{eq:fuse}, LOCO-CV $\alpha$).

\begin{table}[t]
\centering\small
\caption{LoCoMo: BM25\,\oxor{}\,late-interaction fusion vs.\ late interaction
alone (dense fixed at max-sim; LOCO-CV $\alpha$). Hit@1; all $p<10^{-4}$.}
\label{tab:fuse}
\begin{tabular}{lccc}
\toprule
encoder & max-sim & +BM25 & $\Delta$ pp [95\% CI] \\
\midrule
gte-base    & 0.519 & 0.691 & $+17.2$ [13.7, 20.6] \\
bge-base    & 0.547 & 0.699 & $+15.2$ [13.0, 17.2] \\
mxbai-large & 0.578 & 0.713 & $+13.5$ [10.3, 16.7] \\
e5-base-v2  & 0.580 & 0.702 & $+12.1$ [9.8, 14.8] \\
bge-large   & 0.602 & 0.721 & $+11.9$ [9.5, 14.4] \\
e5-large-v2 & 0.664 & 0.752 & $+8.8$ [6.6, 11.2] \\
\bottomrule
\end{tabular}
\end{table}

Fusion adds significantly over late interaction alone for \emph{every}
encoder. The trend is the \textbf{mirror image} of \S\ref{sec:scale}: the
marginal value of BM25 fusion \emph{shrinks} as the dense encoder strengthens
($+17.2$\,pp for gte-base down to $+8.8$\,pp for e5-large-v2). A stronger dense
model leaves BM25 less to add, yet it still adds $\approx9$\,pp at the top.
Lexical fusion is thus not a crutch for weak encoders only; it is a robust,
free addition across the quality range we tested. Table~\ref{tab:fusionfull}
gives the full fusion metrics; Table~\ref{tab:headline} situates the best
recipe against the standard baselines.

\begin{table*}[t]
\centering\small
\caption{LoCoMo BM25\,\oxor{}\,late-interaction fusion (LOCO-CV $\alpha$),
full metrics, six encoders.}
\label{tab:fusionfull}
\begin{tabular}{lccccc}
\toprule
encoder & Hit@1 & R@3 & R@5 & MRR & NDCG@5 \\
\midrule
gte-base    & 0.691 & 0.806 & 0.862 & 0.788 & 0.783 \\
bge-base    & 0.699 & 0.812 & 0.871 & 0.794 & 0.791 \\
e5-base-v2  & 0.702 & 0.819 & 0.871 & 0.798 & 0.794 \\
mxbai-large & 0.713 & 0.824 & 0.879 & 0.806 & 0.802 \\
bge-large   & 0.721 & 0.833 & 0.884 & 0.812 & 0.809 \\
e5-large-v2 & \textbf{0.752} & \textbf{0.846} & \textbf{0.894} & \textbf{0.835} & \textbf{0.829} \\
\bottomrule
\end{tabular}
\end{table*}

\begin{table}[t]
\centering\small
\caption{LoCoMo reference rows (dense/fusion: e5-large-v2, LOCO-CV $\alpha$).}
\label{tab:headline}
\begin{tabular}{lccc}
\toprule
method & Hit@1 & R@5 & NDCG@5 \\
\midrule
BM25 & 0.640 & 0.833 & 0.746 \\
Dense, mean-pool (early) & 0.427 & 0.713 & 0.573 \\
Dense, max-sim (late) & 0.664 & 0.863 & 0.769 \\
BM25 \oxor{} mean-pool & 0.666 & 0.851 & 0.766 \\
\textbf{BM25 \oxor{} max-sim} & \textbf{0.752} & \textbf{0.894} & \textbf{0.829} \\
\bottomrule
\end{tabular}
\end{table}

The deployable recipe (e5-large-v2, BM25\,\oxor{}\,max-sim) reaches Hit@1
$0.752$, $+11.2$\,pp over BM25 (95\% CI $[+9.4,+13.1]$, $p<10^{-4}$; CIs in
\texttt{tune13.json}). Late interaction also matters \emph{within} fusion:
replacing the early dense term with the late term adds $+8.6$\,pp Hit@1
(e5-large-v2, $p<10^{-4}$), so fusion does not wash out the
interaction-function effect. BM25 is a strong baseline here (Hit@1 $0.640$),
so the result reads as a retrieval-stage improvement over strong lexical
matching, not as ``dense dominates lexical.''

\section{Result 2: A web-search cross-encoder does not help}
\label{sec:rerank}
A reflexive way to improve a retriever is to bolt on a cross-encoder reranker.
We test the most common off-the-shelf choice and find it hurts here. Reranking
the fused top-10 sessions with \texttt{ms-marco-MiniLM-L-6-v2} over (query,
best-turn) pairs---on the bge-base max-sim fusion at a fixed $\alpha{=}0.6$---
changes Hit@1 from $0.701$ to $0.633$, a drop of $-6.88$\,pp with 95\% CI
$[-9.34,-4.34]$ that excludes zero (Table~\ref{tab:rerank}).\footnote{The
one-sided bootstrap $p$ tests \emph{improvement} (CE $>$ fusion) and returns
$p{=}1.0$: zero bootstrap support for the reranker helping; equivalently, the
CI lies entirely below 0.} The MS-MARCO cross-encoder is trained on web-search
queries; LoCoMo's conversational, often inferential queries are out of its
distribution, so it reorders a good list into a worse one.

\begin{table}[t]
\centering\small
\caption{Scoped cross-encoder reranking (bge-base max-sim fusion, $\alpha{=}0.6$,
rerank fused top-10).}
\label{tab:rerank}
\begin{tabular}{lcc}
\toprule
condition & Hit@1 & change \\
\midrule
fused first stage & 0.701 & --- \\
+ cross-encoder rerank & 0.633 & $-6.88$ pp \\
\bottomrule
\end{tabular}
\end{table}

This is a \emph{scoped negative result}, not a claim that reranking is
useless: we tested one off-the-shelf web-search cross-encoder, on the bge-base
fusion at a fixed $\alpha{=}0.6$, not the e5-large-v2 headline pipeline, and we
did not try conversational or instruction-tuned rerankers, which might help.
The takeaway is narrow but practical: the default ``retrieve-then-rerank''
move, with the most common cross-encoder, degrades a fused conversational
ranking that already combines lexical and turn-level dense evidence---so a
reranker should be \emph{validated} on the target distribution, not assumed.
(It is also strictly more expensive: a transformer forward pass per candidate,
versus arithmetic on already-computed scores for fusion.)

\section{Result 3: Pooling-Operator Ablation}
\label{sec:ablation}
``Late interaction'' is a family of aggregators, not a single operator.
Table~\ref{tab:pooling} sweeps four operators (dense-only Hit@1): mean (early),
max-sim, top-3, and smooth-max log-sum-exp ($\beta{=}10$).

\begin{table}[t]
\centering\small
\caption{LoCoMo dense-only Hit@1 by pooling operator. Top-3 tracks max-sim,
but smooth-max (lse-10) collapses for half the encoders.}
\label{tab:pooling}
\begin{tabular}{lcccc}
\toprule
encoder & mean & max-sim & top-3 & lse-10 \\
\midrule
gte-base    & 0.384 & 0.519 & 0.517 & 0.126 \\
bge-base    & 0.386 & 0.547 & 0.552 & 0.340 \\
e5-base-v2  & 0.407 & 0.580 & 0.576 & 0.126 \\
mxbai-large & 0.405 & 0.578 & 0.572 & 0.401 \\
bge-large   & 0.403 & 0.602 & 0.590 & 0.340 \\
e5-large-v2 & 0.427 & 0.664 & 0.637 & 0.127 \\
\bottomrule
\end{tabular}
\end{table}

Two observations. First, \textbf{top-3 tracks max-sim} (within $\sim$1--3\,pp
everywhere): aggregating the few best turns is as good as taking the single
best, so the result is not an artifact of the exact $\max$. Second,
\textbf{smooth-max collapses} for e5-base-v2, gte-base, and e5-large-v2 (to
$\approx0.13$ Hit@1, below even mean-pooling) while surviving for bge and
mxbai. Log-sum-exp is sensitive to the absolute scale of the cosine
distribution, which differs across encoders; a fixed $\beta$ that is
reasonable for one encoder saturates or vanishes for another. The practical
lesson is that the \emph{robust} late-interaction operators are the
scale-free ones (max, top-$k$); the family is not uniformly safe, which
matters for anyone porting the recipe across encoders. Notably, BM25 fusion
masks the collapse---fused lse-10 recovers to $0.65$--$0.67$ Hit@1---because
the lexical term carries the ranking when the dense term degenerates, a
further argument for fusion as a robustness mechanism.

\section{Result 4: Cross-Corpus Boundary (LongMemEval-S)}
\label{sec:cross}
LongMemEval-S is the opposite lexical regime: heavy query--answer lexical
overlap, so BM25 is very strong. Table~\ref{tab:lme} reports the same control
with all-MiniLM-L6-v2 on the 150-question challenge subset, including top-3
and RRF.

\begin{table}[t]
\centering\small
\caption{LongMemEval-S challenge subset (all-MiniLM-L6-v2, $n{=}150$).}
\label{tab:lme}
\begin{tabular}{lccc}
\toprule
method & R@1 & R@5 & NDCG@5 \\
\midrule
BM25 & 0.589 & 0.948 & 0.916 \\
Dense, early (mean) & 0.545 & 0.920 & 0.876 \\
Dense, late (max-sim) & 0.568 & 0.953 & 0.906 \\
Dense, late (top-3) & 0.569 & 0.951 & 0.908 \\
RRF(BM25, late) & 0.594 & 0.956 & 0.932 \\
\textbf{BM25 \oxor{} late} & \textbf{0.594} & \textbf{0.959} & \textbf{0.930} \\
\bottomrule
\end{tabular}
\end{table}

Two readings. \textbf{(1)~The direction holds but is not significant
here.} Dense late beats dense early by $+4.67$\,pp R@1, but the 95\% CI is
$[-1.33,+10.67]$ ($p{=}0.071$)---directional, not significant at $.05$. It is
nonetheless \emph{consistent} with the encoder-scaling trend by ordering:
MiniLM-22M's $+4.67$\,pp gap is smaller than every (larger) LoCoMo encoder's
gap. \textbf{(2)~The net win over BM25 does not transfer.} The fusion margin
over BM25 is $+0.67$\,pp R@1 (95\% CI $[-2.67,+4.00]$, $p{=}0.43$), not
significant; RRF and weighted fusion are indistinguishable here. Every method
exceeds the published Stella-1.5B session retriever (R@5 $0.732$;
\citealp{longmemeval}), but that is an uncontrolled cross-paper comparison and
reflects \textbf{BM25's} lexical dominance, not late interaction's.
LongMemEval-S is therefore an honest \textbf{boundary}: the mechanism is real
but its payoff depends on how much the corpus rewards semantic over lexical
matching---large on LoCoMo, negligible where BM25 saturates. The
strong-encoder field-metric point is missing (bge-base stalls embedding
LongMemEval's thousands-of-turn haystacks on CPU); closing it is the main
experiment for a camera-ready version.

\section{Analysis: When and Why Fusion Helps}
\label{sec:analysis}
The aggregate ``fusion wins'' result hides a more useful story about
\emph{which} queries each signal answers, why the gap depends on session
length, and how robust the recipe is to its one hyperparameter.

\begin{figure}[t]
\centering
\includegraphics[width=\linewidth]{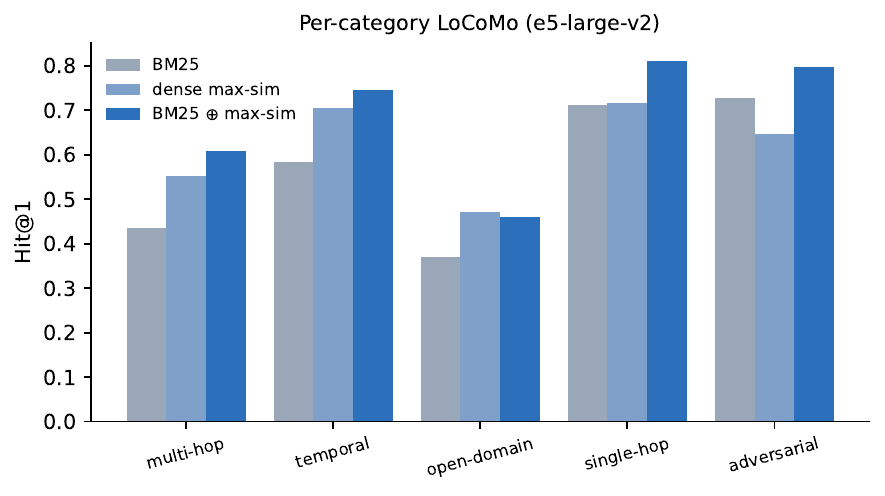}
\caption{Per-category LoCoMo Hit@1 (e5-large-v2). Dense max-sim wins on
multi-hop/temporal, ties on single-hop, and \emph{loses} to BM25 on
adversarial; fusion wins in every category.}
\label{fig:category}
\end{figure}

\paragraph{A division of labor across question types.}
Table~\ref{tab:category} and Fig.~\ref{fig:category} break LoCoMo Hit@1 down by
question category (e5-large-v2). Late-interaction dense retrieval is strongest
exactly where lexical matching is weakest: it beats BM25 by $+11.8$\,pp on
multi-hop and $+12.1$\,pp on temporal questions---the categories that require
linking or ordering evidence. But on \emph{adversarial} questions, designed to
bait surface matching, dense \emph{underperforms} BM25 by $8.0$\,pp. Fusion is
the hedge: it beats BM25 in every category ($+7.0$ to $+17.4$\,pp) and beats
dense in four of five. The lone exception is open-domain ($n{=}89$), where BM25
is so weak ($0.371$) that adding it slightly drags fusion below pure dense
($0.461$ vs.\ $0.472$); we note it. The practical
reading is that the reflexive ``pick the single best retriever'' choice is
wrong on a large fraction of queries---no single retriever dominates across
categories, and fusion captures whichever signal a query needs.

\begin{table}[t]
\centering\small
\caption{LoCoMo Hit@1 by question category (e5-large-v2). Dense helps on
reasoning categories but hurts on adversarial; fusion beats BM25 everywhere.}
\label{tab:category}
\begin{tabular}{lccccc}
\toprule
category & $n$ & BM25 & dense & fusion & {\small fus$-$bm} \\
\midrule
multi-hop    & 281 & 0.434 & 0.552 & \textbf{0.609} & $+17.4$ \\
temporal     & 321 & 0.583 & 0.704 & \textbf{0.745} & $+16.2$ \\
single-hop   & 841 & 0.712 & 0.716 & \textbf{0.810} & $+9.8$ \\
open-domain  &  89 & 0.371 & \textbf{0.472} & 0.461 & $+9.0$ \\
adversarial  & 446 & 0.726 & 0.646 & \textbf{0.796} & $+7.0$ \\
\bottomrule
\end{tabular}
\end{table}

\paragraph{The gap grows with session length.}
Table~\ref{tab:length} buckets the late$-$early Hit@1 gap by the number of
turns in the gold session. For every encoder the gap \emph{increases
monotonically with length}: e5-large-v2 rises from $+19.4$\,pp at 8--15 turns
to $+27.1$\,pp at 26+. This is direct evidence for the dilution mechanism---a
longer session gives mean-pooling more turns to average the answer-bearing turn
against---and is a second axis, orthogonal to encoder quality
(\S\ref{sec:scale}), along which the effect strengthens.

\begin{table}[t]
\centering\small
\caption{LoCoMo late$-$early Hit@1 gap (pp) by gold-session length. The gap
grows with length for all six encoders. Bucket sizes $n=186/1123/669$.}
\label{tab:length}
\begin{tabular}{lccc}
\toprule
encoder & 8--15 & 16--25 & 26+ \\
\midrule
gte-base    & $+12.4$ & $+13.1$ & $+14.5$ \\
bge-base    & $+14.5$ & $+14.9$ & $+18.5$ \\
e5-base-v2  & $+9.1$  & $+16.2$ & $+21.4$ \\
mxbai-large & $+9.1$  & $+16.9$ & $+20.2$ \\
bge-large   & $+11.8$ & $+19.6$ & $+22.6$ \\
e5-large-v2 & $+19.4$ & $+22.4$ & $+27.1$ \\
\bottomrule
\end{tabular}
\end{table}

\begin{figure}[t]
\centering
\includegraphics[width=0.92\linewidth]{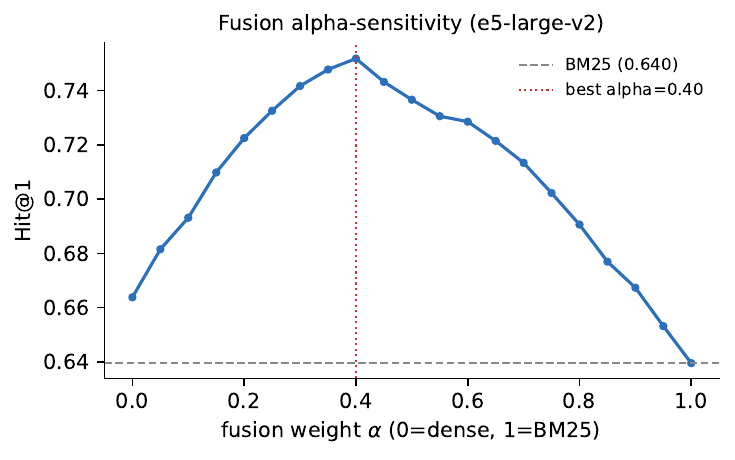}
\caption{Fusion is robust to its weight: LoCoMo Hit@1 vs.\ global $\alpha$
(e5-large-v2). A broad plateau peaks at $\alpha{=}0.40$; LOCO-CV selects this
optimum. $\alpha{=}0$ is pure dense, $\alpha{=}1$ pure BM25.}
\label{fig:alpha}
\end{figure}

\paragraph{Fusion weight and combiner.}
The fusion weight is not knife-edge. Figure~\ref{fig:alpha} sweeps the global
weight $\alpha$ (Eq.~\ref{eq:fuse}): Hit@1 exceeds $0.73$ across the broad range
$\alpha\in[0.25,0.50]$ and peaks at $\alpha{=}0.40$ ($0.752$), well above both
pure dense ($\alpha{=}0$, $0.664$) and pure BM25 ($\alpha{=}1$, $0.640$).
Leave-one-conversation-out CV selects exactly this optimum, confirming the
recipe is not overfit to a hand-tuned weight. On combiners, reciprocal rank
fusion (RRF) reaches $0.718$ Hit@1---above BM25 and dense, but $3.4$\,pp below
the $z$-normalized weighted fusion ($0.752$). Weighted score-level fusion with
a single global $\alpha$ is the better, simpler choice here.

\section{Discussion}
\label{sec:discussion}
\paragraph{Reconciling the granularity literature.}
The early/late distinction explains why the literature disagrees about ``turn
vs.\ session.'' A reader-centric pipeline that feeds whole-session
\emph{text} to an LLM never pools embeddings and reports ``session is fine'';
a dense pipeline that \emph{embeds} a pooled session pays the
early-interaction penalty and reports ``session dense is poor.'' Both are
right about their own configuration; the hidden variable is the interaction
function, not the unit. Our control makes this explicit by holding the unit
fixed---an argument also advanced by Nano-Memory \citep{nanomemory} from the
sparsity side.

\paragraph{When does fusion help?}
The two scaling trends (\S\ref{sec:scale}, \S\ref{sec:fuse}) compose into a
simple picture: the dense late-interaction term contributes more as the
encoder improves, while the lexical term contributes more when the dense term
is weak or when the corpus is lexical (LongMemEval). Fusion is the hedge that
captures whichever signal is available---and, as the lse-10 collapse shows
(\S\ref{sec:ablation}), it also rescues a degenerate dense operator.

\paragraph{Practical recommendation.}
For training-free, CPU-only conversational-memory retrieval: (i) score
sessions by turn-level late interaction (max-sim or top-$k$, not smooth-max);
(ii) fuse with BM25 at the score level under a single LOCO-CV weight;
(iii) prefer a stronger encoder, which widens the dense advantage; and
(iv) do not add a cross-encoder reranker without validating it on the target
query distribution.

\section{Limitations and Threats to Validity}
\label{sec:limits}
\textbf{Concurrency.} Nano-Memory \citep{nanomemory} appeared $\sim$April 2026;
the late-interaction retriever and the max-vs-mean insight are theirs, and we
build around them. \textbf{Retrieval-stage only.} We use no LLM reader and make
no end-to-end QA claim; bridging to answer accuracy is the natural next step.
\textbf{Single benchmark; few clusters.} All positive results are on LoCoMo,
and its conversation-cluster bootstrap resamples only \emph{10} conversations:
the effective number of independent units is small, so the LoCoMo $p$-values,
though tiny, rest on a narrow base. A third benchmark would reduce this
fragility. \textbf{Capacity trend, not a law.} The late$-$early gap grows with
encoder capacity only as a soft, non-monotone trend over six encoders; we
deliberately avoid a gap-vs-own-score correlation as near-circular
(\S\ref{sec:scale}). \textbf{Core novelty is incremental.} Hybrid
lexical--dense fusion is well established; our contribution is the controlled
demonstration that it adds over a turn-level late-interaction dense arm
specifically, with the accompanying ablations---not a new retrieval mechanism.
\textbf{Cross-corpus point.} LongMemEval-S uses a 22M encoder at $n{=}150$; the
strong-encoder field-metric run is CPU-prohibitive (we confirmed this
empirically---e5-base-v2 stalled embedding the thousands-of-turn haystacks and
made no progress past the first example, the same failure mode as bge-base), so
the boundary claim is honest but underpowered and awaits a GPU or
chunked-embedding run. \textbf{Reranking scope.} The
reranking result is one cross-encoder in one (bge-base, $\alpha{=}0.6$)
condition, not an exhaustive reranker study. \textbf{Prior-art currency.} This is a fast-moving area, with several relevant
papers appearing within six months of this work. We therefore treat the
related-work section as a snapshot and recommend updating comparisons as new
conversational-memory systems appear.

\section{Conclusion}
For conversational-memory retrieval, the strongest simple recipe in our
experiments is neither dense-only turn isolation nor cross-encoder
reranking---it is score-level fusion of BM25 with turn-level late interaction.
Fusion adds significantly over late interaction alone across six encoders
($+8.8$ to $+17.2$\,pp Hit@1, all $p<10^{-4}$), the best version improves over
a strong BM25 baseline by $+11.2$\,pp Hit@1, and the recipe is training-free
and CPU-only. It has clear boundaries: an off-the-shelf web-search cross-encoder
reranker hurts, a smooth-max pooling operator collapses for half the encoders,
and on a lexical-overlap corpus the net fusion gain over BM25 is not
significant. We credit turn isolation to Nano-Memory and contribute the
controlled account of what to add around it: fuse lexical and late-interaction
dense scores, choose a scale-free late operator (max-sim or top-$k$), and
validate any reranker on the target distribution rather than assume it helps. This is not a memory \emph{system}
and makes no end-to-end-SOTA claim; it is best read as a strong, training-free,
CPU-only retrieval baseline that more elaborate memory architectures (graph
memory, learned segmentation, query-driven pruning) should be required to beat
before their added complexity is justified. Combining the fusion recipe with
Nano-Memory's query-driven pruning is the natural next step.

\paragraph{Reproducibility.}
All results are CPU-only and run from cached embeddings. Code, receipts (the
exact JSON/markdown outputs behind every table), and the figure script are at
\url{https://github.com/Chrislysen/opsem}: \texttt{tune13\_interaction.py},
\texttt{tune13b\_fusion\_vs\_late.py}, \texttt{tune10\_rerank.py},
\texttt{lme\_interaction.py}, \texttt{make\_figures.py}.

\bibliographystyle{unsrtnat}
\bibliography{references}

@inproceedings{locomo,
  title     = {Evaluating Very Long-Term Conversational Memory of {LLM} Agents},
  author    = {Maharana, Adyasha and Lee, Dong-Ho and Tulyakov, Sergey and Bansal, Mohit and Barbieri, Francesco and Fang, Yuwei},
  booktitle = {Proceedings of the Annual Meeting of the Association for Computational Linguistics (ACL)},
  year      = {2024},
  note      = {arXiv:2402.17753}
}

@inproceedings{longmemeval,
  title     = {{LongMemEval}: Benchmarking Chat Assistants on Long-Term Interactive Memory},
  author    = {Wu, Di and Wang, Hongwei and Yu, Wenhao and Zhang, Yuwei and Chang, Kai-Wei and Yu, Dong},
  booktitle = {International Conference on Learning Representations (ICLR)},
  year      = {2025},
  note      = {arXiv:2410.10813}
}

@inproceedings{secom,
  title     = {{SeCom}: On Memory Construction and Retrieval for Personalized Conversational Agents},
  author    = {Pan, Zhuoshi and Wu, Qianhui and Jiang, Huiqiang and Luo, Xufang and Cheng, Hao and Li, Dongsheng and Yang, Yuqing and Lin, Chin-Yew and Zhao, H. Vicky and Qiu, Lili and Gao, Jianfeng},
  booktitle = {International Conference on Learning Representations (ICLR)},
  year      = {2025},
  note      = {arXiv:2502.05589}
}

@misc{nanomemory,
  title         = {Back to Basics: Let Conversational Agents Remember with Just Retrieval and Generation},
  author        = {Wu, Yuqian and Chen, Wei and Huang, Zhengjun and Chen, Junle and Liu, Qingxiang and Wang, Kai and Zhou, Xiaofang and Liang, Yuxuan},
  year          = {2026},
  eprint        = {2604.11628},
  archivePrefix = {arXiv},
  note          = {Nano-Memory. \url{https://github.com/yuqian2003/Nano-Memory}}
}

@inproceedings{colbert,
  title     = {{ColBERT}: Efficient and Effective Passage Search via Contextualized Late Interaction over {BERT}},
  author    = {Khattab, Omar and Zaharia, Matei},
  booktitle = {Proceedings of the International ACM SIGIR Conference on Research and Development in Information Retrieval (SIGIR)},
  year      = {2020},
  note      = {arXiv:2004.12832}
}

@article{bm25,
  title     = {The Probabilistic Relevance Framework: {BM25} and Beyond},
  author    = {Robertson, Stephen and Zaragoza, Hugo},
  journal   = {Foundations and Trends in Information Retrieval},
  volume    = {3},
  number    = {4},
  pages     = {333--389},
  year      = {2009}
}

@inproceedings{rrf,
  title     = {Reciprocal Rank Fusion Outperforms Condorcet and Individual Rank Learning Methods},
  author    = {Cormack, Gordon V. and Clarke, Charles L. A. and B{\"u}ttcher, Stefan},
  booktitle = {Proceedings of the International ACM SIGIR Conference on Research and Development in Information Retrieval (SIGIR)},
  year      = {2009}
}

@misc{memgpt,
  title         = {{MemGPT}: Towards {LLMs} as Operating Systems},
  author        = {Packer, Charles and Wooders, Sarah and Lin, Kevin and Fang, Vivian and Patil, Shishir G. and Stoica, Ion and Gonzalez, Joseph E.},
  year          = {2023},
  eprint        = {2310.08560},
  archivePrefix = {arXiv}
}

@misc{zep,
  title         = {Zep: A Temporal Knowledge Graph Architecture for Agent Memory},
  author        = {Rasmussen, Preston and Paliychuk, Pavlo and Beauvais, Travis and Ryan, Jack and Chalef, Daniel},
  year          = {2025},
  eprint        = {2501.13956},
  archivePrefix = {arXiv}
}

@misc{e5,
  title         = {Text Embeddings by Weakly-Supervised Contrastive Pre-training},
  author        = {Wang, Liang and Yang, Nan and Huang, Xiaolong and Jiao, Binxing and Yang, Linjun and Jiang, Daxin and Majumder, Rangan and Wei, Furu},
  year          = {2022},
  eprint        = {2212.03533},
  archivePrefix = {arXiv},
  note          = {E5}
}

@misc{bge,
  title         = {{C-Pack}: Packed Resources for General Chinese Embeddings},
  author        = {Xiao, Shitao and Liu, Zheng and Zhang, Peitian and Muennighoff, Niklas},
  year          = {2023},
  eprint        = {2309.07597},
  archivePrefix = {arXiv},
  note          = {BGE}
}

@misc{gte,
  title         = {Towards General Text Embeddings with Multi-stage Contrastive Learning},
  author        = {Li, Zehan and Zhang, Xin and Zhang, Yanzhao and Long, Dingkun and Xie, Pengjun and Zhang, Meishan},
  year          = {2023},
  eprint        = {2308.03281},
  archivePrefix = {arXiv},
  note          = {GTE}
}

@misc{mxbai,
  title        = {Open Source Strikes Bread --- New Fluffy Embeddings Model ({mxbai-embed-large-v1})},
  author       = {Lee, Sean and Shakir, Aamir and Koenig, Darius and Lipp, Julius},
  year         = {2024},
  howpublished = {Mixedbread AI},
  note         = {Model card}
}

@inproceedings{sbert,
  title     = {{Sentence-BERT}: Sentence Embeddings using Siamese {BERT}-Networks},
  author    = {Reimers, Nils and Gurevych, Iryna},
  booktitle = {Proceedings of the Conference on Empirical Methods in Natural Language Processing (EMNLP)},
  year      = {2019},
  note      = {arXiv:1908.10084; all-MiniLM models}
}

\end{document}